\documentclass[10pt,twocolumn,letterpaper]{article}

\usepackage{iccv}              

\definecolor{iccvblue}{rgb}{0.21,0.49,0.74}
\usepackage[pagebackref,breaklinks,colorlinks,allcolors=iccvblue]{hyperref}
\usepackage{multirow}
\usepackage{comment}
\def\paperID{14} 
\def\confName{ICCV}
\def\confYear{2025}

\title{iDETEX: Empowering MLLMs for Intelligent DETailed EXplainable IQA}

\author{Zhaoran Zhao\thanks{Equal contribution.}
\hspace{.1in}
Xinli Yue\footnotemark[1]
\hspace{.1in}
Jianhui Sun
\hspace{.1in}
Yuhao Xie \\
Tao Shao
\hspace{.1in}
Liangchao Yao
\hspace{.1in}
FAN XIA
\hspace{.1in}
Yuetang Deng\thanks{Corresponding author.} \\
Tencent, WeChat\\
\tt\small\{xinliyue,nimosun,yohoxie,taoshao,clarkyao,frankxia,yuetangdeng\}@tencent.com \\
}

\begin{document}
\maketitle
\begin{abstract}

Image Quality Assessment (IQA) has progressed from scalar quality prediction to more interpretable, human-aligned evaluation paradigms. In this work, we address the emerging challenge of detailed and explainable IQA by proposing iDETEX—a unified multimodal large language model (MLLM) capable of simultaneously performing three key tasks: quality grounding, perception, and description. To facilitate efficient and generalizable training across these heterogeneous subtasks, we design a suite of task-specific offline augmentation modules and a data mixing strategy. These are further complemented by online enhancement strategies to fully exploit multi-sourced supervision. We validate our approach on the large-scale ViDA-UGC benchmark, where iDETEX achieves state-of-the-art performance across all subtasks. Our model ranks first in the ICCV MIPI 2025 Detailed Image Quality Assessment Challenge, demonstrating its effectiveness and robustness in delivering accurate and interpretable quality assessments.

\end{abstract}    

\section{Introduction}

Image Quality Assessment (IQA)~\cite{wang2004image_iqa_intro} plays a vital role in a wide range of computer vision applications~\cite{chow2016review_iqa_app,ying2020patches_iqa_app,wu2023assessor360_iqa_app,fang2020perceptual_iqa_app}, aiming to evaluate perceptual quality in a manner consistent with the Human Visual System (HVS).
Traditional IQA methods typically rely on scalar quality scores to quantify image quality~\cite{shin2024blind_learning_base_iqa,liu2017rankiqa_learning_base_iqa,kang2014convolutional_learning_base_iqa}. 
While effective in some cases, these approaches generally lack transparency and interpretability.

With the rapid advancement of Multimodal Large Language Models (MLLMs)~\cite{qin2024multilingual_mllm_survey,caffagni2024revolution_mllm_survey}, the field is undergoing a paradigm shift from black-box quality scoring to explainable image quality assessment. In this new paradigm, models are expected not only to output quality scores but also to identify distortion regions, perceive visual attributes, and generate human-readable descriptions, thereby aligning more closely with human perception of image quality.

\begin{figure*}[t]
    \centering
    \includegraphics[width=1.\textwidth]{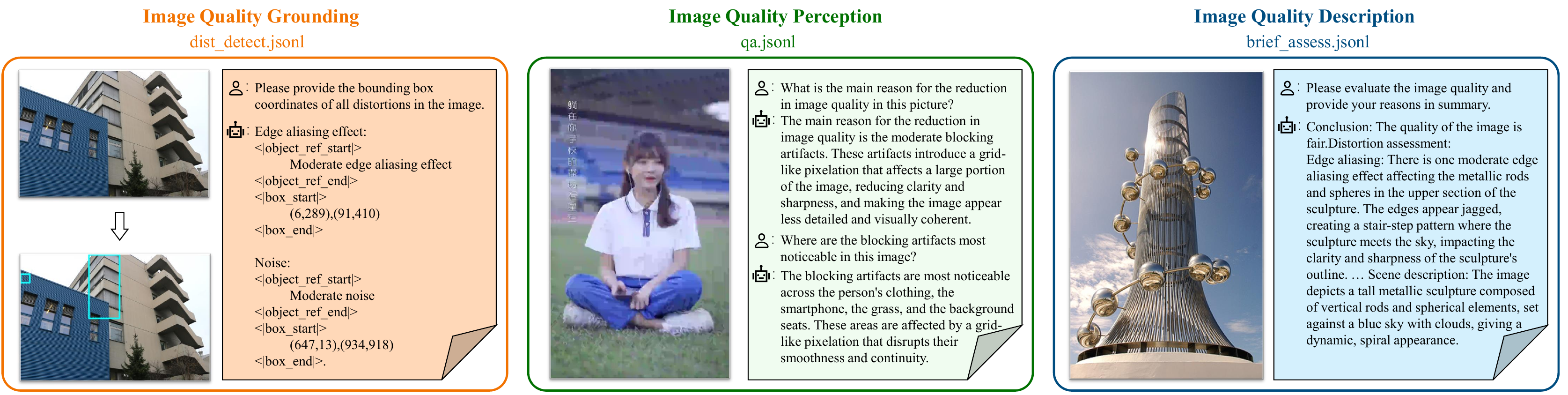}
    \caption{Illustration of the input and output formats for the three core tasks for explainable IQA in ViDA-UGC~\cite{DYEvalab2025ViDAMIPI}, using representative examples. From left to right: Image Quality Grounding, Image Quality Perception, and Image Quality Description.}
    \label{fig:intro}
\end{figure*}

Specifically, as shown in Fig.~\ref{fig:intro}, explainable IQA typically consists of three tightly coupled subtasks: \textbf{quality grounding}, \textbf{quality perception}, and \textbf{quality description}. 
The first task focuses on pinpointing the spatial locations of distortions, aligning HVS sensitivity to local degradations. Image quality perception delves into analyzing fine-grained visual attributes—both within the distorted areas and across the entire image—capturing a broader range of perceptual cues similar to those processed by the HVS. 
Image quality description synthesizes the spatial and perceptual insights into structured, interpretable feedback, forming a causal reasoning chain that mirrors human judgment beyond basic visual recognition. 

Although the above tasks are fundamental to explainable IQA, training MLLMs that can simultaneously perform all of them remains highly challenging:
\begin{itemize}
\item The multi-task learning process is inherently complex and demands carefully designed training strategies to ensure effective optimization across tasks.
\item Existing explainable IQA datasets are limited in scale and expensive to annotate, making it essential to maximize data efficiency through augmentation and task-specific supervision.
\end{itemize}

To address the aforementioned challenges, we propose a series of task-specific offline data augmentation mechanisms with each subtask equipped with a dedicated augmentation module to fully exploit available supervision and enhance learning efficiency. Building on this foundation, we design a data mixing strategy to effectively integrate the above multi-sourced data for MLLM training.

With the support of offline augmented data and complementary online augmentation strategies, we successfully train iDETEX — a unified Multimodal Large Language Model network for Intelligent DETailed and EXplainable IQA, capable of simultaneously performing the three core tasks: quality grounding, quality perception, and quality description.

We train and evaluate our method on ViDA-UGC~\cite{DYEvalab2025ViDAMIPI}, a large-scale, richly annotated dataset specifically designed for fine-grained and explainable IQA provided in MIPI 2025 Detailed Image Quality Assessment Challenge. 
Experimental results demonstrate that our model, iDETEX, delivers accurate, robust, and interpretable performance across all subtasks. Our method ranks first on the official leaderboard of the challenge.

Our main contributions are summarized as follows:

\begin{itemize}
    \item Task-specific augmentation for fine-tuning. We develop three dedicated augmentation modules tailored to the unique requirements of each subtask, along with a data mixing strategy that enhances spatial localization, perceptual sensitivity, and causal reasoning ability. These augmentation methods enable MLLMs to be trained efficiently and with strong generalization capability across the three core tasks.

    \item iDETEX: A unified MLLM-based network for explainable IQA. We present iDETEX, a multimodal large language model designed to perform three core explainable IQA tasks—quality grounding, perception, and description—in a unified and interpretable manner.

    \item State-of-the-art results on ViDA-UGC-Bench. Our model achieves top performance on the ViDA-UGC-Bench benchmark for detailed IQA. iDETEX ranks first on the official leaderboard of the Detailed Image Quality Assessment Challenge.
\end{itemize}

\section{Related Work}
IQA models can be broadly categorized into convolutional neural network (CNN)-based and Transformer-based approaches. CNN-based models~\cite{cnn-iqa1-kang2014convolutional,cnn-iqa2-kang2015simultaneous,cnn-iqa3-su2020blindly} excel at capturing local features, achieving strong performance in quality prediction by combining feature learning with regression. These methods have progressively incorporated multi-task learning and multi-scale feature fusion to further enhance representation capabilities. In contrast, Transformer-based models~\cite{yang2022maniqa_iqa_trans} address the limitations of CNNs in modeling non-local dependencies by leveraging self-attention mechanisms to strengthen global interactions among image regions, thereby improving quality perception. Some recent works~\cite{qin2023data_iqa_both,golestaneh2022no_iqa_both} integrate the strengths of both CNNs and Transformers to reduce local biases and augment global context modeling, leading to more accurate quality assessment. 

Recent progress in Multimodal Large Language Models (MLLMs) has shown promising results in unifying vision and language modalities by coupling visual encoders with large-scale language models~\cite{qin2024multilingual_mllm_survey,caffagni2024revolution_mllm_survey,li2024_survey_mllm}. These models excel at various high-level vision-language tasks, such as image captioning and visual question answering. More recently, researchers have begun to explore their applicability to low-level perceptual tasks like image quality assessment (IQA), extending their utility beyond traditional semantic understanding. To facilitate this transition, several IQA-focused benchmarks have been proposed to systematically evaluate the descriptive, comparative, and evaluative abilities of MLLMs under real-world degradations. Q-Bench~\cite{wu2023q_qbench} and DepictQA~\cite{ying2020patches_DepictQA}, for example, offer structured evaluation sets that capture human-perceived image quality. To strengthen perceptual alignment, Q-Instruct~\cite{wu2024q_q_instruct} and Co-Instruct~\cite{wu2024towards_co_instruct} introduce large-scale instruction tuning tailored for quality-aware understanding, while Q-Align~\cite{wu2023q_q_align} employs a discrete scoring mechanism to improve output consistency with subjective scores. Despite these advancements, current MLLM-based IQA frameworks still struggle with fine-grained tasks such as local distortion identification and detailed quality perception. 
\section{Method}

\begin{figure*}[t]
    \centering
    \includegraphics[width=1.\textwidth]{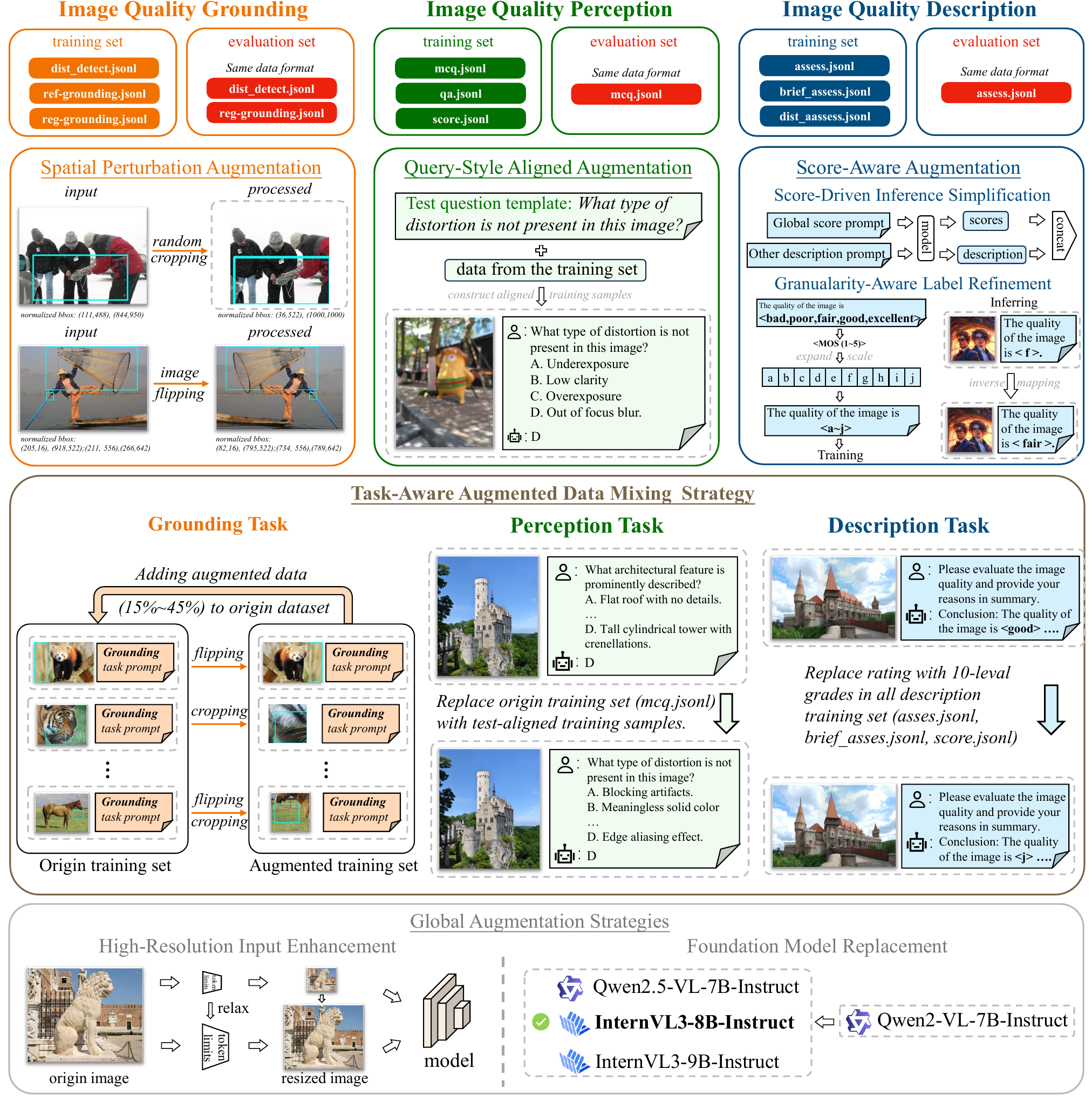}
    \caption{Overview of our training paradigm for iDETEX. Three task-specific augmentation strategies are proposed—Spatial Perturbation Augmentation, Query-Style Aligned Augmentation, and Score-Aware Augmentation—designed for the grounding, perception, and description tasks, respectively. Based on these, we introduce a Task-Aware Augmented Data Mixing Strategy that enables multi-source data to effectively participate in MLLM fine-tuning. These offline augmentations, together with an online High-Resolution Input Enhancement strategy, form a unified fine-tuning pipeline. The augmented data is used to fine-tune a replaced base MLLM, resulting in iDETEX, a model capable of delivering accurate, robust, and interpretable quality assessments across diverse distortion scenarios.
}
    \label{fig:full_net}
\end{figure*}

As illustrated in Fig.~\ref{fig:full_net}, to enable MLLMs to effectively perform the three core tasks of Image Quality Grounding, Perception, and Description for high-quality detailed IQA, we design three task-specific data augmentation strategies: Spatial Perturbation Augmentation~\ref{sec_for_grounding}, Query-Style Aligned Augmentation~\ref{sec_for_Perception}, and Score-Aware Augmentation~\ref{sec_for_Description}. To further integrate multi-source supervision, we introduce a Task-Aware Augmented Data Mixing Strategy~\ref{sec_mixing}, which allows heterogeneous data to jointly contribute to the training of the multimodal model. These offline augmentation strategies are complemented by an online High-Resolution Input Enhancement technique, together forming a unified fine-tuning framework. The augmented data is used to fine-tune a pretrained base MLLM, resulting in iDETEX, a unified model capable of generating accurate, robust, and interpretable assessments across a wide range of distortion types.

\subsection{Spatial Perturbation Augmentation for Grounding}
\label{sec_for_grounding}

To improve the robustness and generalization of the Image Quality Grounding task, which requires precise localization of distortion regions, we introduce a set of spatial perturbation strategies—random cropping and horizontal flipping. These augmentations enrich the spatial distribution of training samples and enhance the model’s ability to detect diverse and position-sensitive distortions.

\subsubsection{Random Cropping}

We apply random cropping by extracting a sub-region covering certain proportions of the original image area. Formally, given an input image \( I \in \mathbb{R}^{H \times W \times 3} \), we randomly sample a crop region \( I' \in \mathbb{R}^{\alpha H \times \alpha W \times 3} \), where \(\alpha \in (0, 1]\) is the cropping ratio. This operation helps the model recognize local distortions even with incomplete global context, enhancing its ability to perceive localized quality degradations. It also mitigates spatial bias by disrupting fixed distortion patterns, encouraging the model to focus on content rather than absolute position or size.

\begin{equation}
I' = \text{Crop}(I; \alpha), \quad \alpha \in (0, 1].
\label{eq:random_cropping}
\end{equation}

\subsubsection{Horizontal Flipping}

Horizontal flipping mirrors the image along the vertical axis. Formally, for an input image \( I \), the horizontally flipped image \( I^{f} \) is defined pixel-wise as:

\begin{equation}
I^{f}(x, y) = I(W - x - 1, y),
\label{eq:horizontal_flipping}
\end{equation}
where \( W \) is the width of the image, and \((x, y)\) are the pixel coordinates. This transformation enhances the model's ability to generalize across symmetric spatial layouts and directional artifacts common in real-world distortions.

\subsubsection{Adjustment of Bounding Box Coordinates}

In addition to the image augmentations, the coordinates of the distortion bounding boxes and candidate boxes also need to be adjusted accordingly. Specifically, when applying random cropping, the region of the distortion within the image is cropped, and the bounding box coordinates must be rescaled to match the new crop size. The coordinates of the bounding boxes are scaled by the same factor $\alpha$ used for cropping, ensuring that they remain relative to the cropped sub-region. Similarly, when performing horizontal flipping, the bounding box coordinates must be updated to reflect the mirrored image. This is achieved by adjusting the horizontal position of the bounding box, effectively inverting its x-coordinate with respect to the image’s width. These transformations ensure that the bounding boxes remain accurate and aligned with the augmented image, preserving the integrity of the localization task.

Together, these perturbations act as implicit regularizers that promote spatial invariance and diversity, improving the accuracy and robustness of distortion localization.

\subsection{Query-Style Aligned Augmentation for Perception}
\label{sec_for_Perception}

The image quality perception task focuses on analyzing low-level visual attributes across both distorted regions and the overall image. 
It typically requires the model to select the most appropriate description from multiple semantically similar candidates. 
Success in this task depends not only on visual understanding but also on the model’s ability to interpret the structure and phrasing of the question.

To improve model performance in this setting, we propose a query-style aligned augmentation strategy.
Specifically, we construct training samples that closely match the question format used in the test set, ensuring that the model is exposed to consistent query styles during training. 
This alignment reduces distributional discrepancies between training and testing and helps the model better understand the intent behind each question, ultimately improving its answer selection accuracy.

{
We clarify that our modification was stylistic, not semantic. We only rephrased questions to match the test set’s format, while the full diversity of attributes in the original dataset was preserved. This strategy does not harm generalization; instead, it allows the model to apply its knowledge more robustly by removing confusion from varying query structures.
}

\subsection{Score-Aware Augmentation for Description}
\label{sec_for_Description}

The image quality description task aims to establish a causal reasoning chain that connects low-level distortion attributes with high-level perceptual judgments, ultimately producing interpretable and structured quality feedback. 

To improve the model’s reasoning ability and scoring accuracy, we propose two complementary strategies under a unified framework called Score-Aware Augmentation:

\subsubsection{Score-Driven Inference Simplification}  
In the original task setting, the model must simultaneously infer distortions, identify the key degradation, and predict the overall score, which can introduce interference between subtasks. To alleviate this, we directly employ the scoring supervision from the perception task during inference, where the model is prompted to predict only the global score. The other two outputs are still derived using the original multi-component prompts. This simplification reduces cognitive load during multi-task learning and improves the consistency and stability of score predictions.

\subsubsection{Granularity-Aware Label Refinement}

\paragraph{Original Method:}
The original IQA task involves assigning an image quality grade based on its Mean Opinion Score (MOS), which is a continuous value ranging from 1 to 5. The MOS score is used to classify the image into one of five discrete quality levels: \textit{bad}, \textit{poor}, \textit{fair}, \textit{good}, and \textit{excellent}. To map the MOS score into these discrete levels, the range [1, 5] is divided into five equal intervals:
\begin{equation}
L_5(s) = l_i \quad \text{if} \quad 1 + \frac{i-1}{5} \times (5 - 1) \leq s < 1 + \frac{i}{5} \times (5 - 1),
\end{equation}
where $\{l_i|_{i=1} ^5\}=\{\text {bad, poor, fair, good, excellent}\}$ and $s$ is the MOS score.

\paragraph{Enhanced Method:}
We introduce finer granularity in the quality assessment by increasing the number of discrete levels. Specifically, we divide the MOS range [1, 5] into more intervals, such as 10, 15, or 20 levels. For instance, if we opt for 10 levels, the MOS range [1, 5] is divided into 10 equal intervals:
\begin{equation}
L_{10}(s) = l_i \quad \text{if} \quad 1 + \frac{i-1}{10} \times (5 - 1) \leq s < 1 + \frac{i}{10} \times (5 - 1),
\end{equation}
where \{$l_i|_{i=1} ^{10}\}=\{\text {a, b, c, d, e, f, g, h, i, j}\}$.

After performing the classification into one of these finer levels, we map the results back to the original 5-level scale for consistency in the evaluation. This mapping follows a predefined scheme, such as:
\begin{equation}
{L}_\text{5} =
\begin{cases}
\text{bad}, & \text{if } L_{10} \in \{a, b\} \\
\text{poor}, & \text{if } L_{10} \in \{c, d\} \\
\text{fair}, & \text{if } L_{10} \in \{e, f\} \\
\text{good}, & \text{if } L_{10} \in \{g, h\} \\
\text{excellent}, & \text{if } L_{10} \in \{i, j\}
\end{cases}.
\end{equation}

{
We clarify that the abstract labels (a-j) are an intermediate, internal representation used only to refine the model's training. For all final results, they are mapped back to the standard 5-level scale, thus maintaining full human interpretability.
}

\subsection{Task-Aware Augmented Data Mixing} 
\label{sec_mixing}
To enhance the synergy and generalization ability in multi-task learning, we propose a Task-Aware Augmented Data Mixing strategy. Specifically, we replace a portion of the original training samples with task-specific augmented data and conduct joint fine-tuning across all subtasks using this enriched dataset. This approach preserves the structural integrity of each task while injecting more discriminative and diverse visual-linguistic signals into the training process. The specific strategies are as follows:

\begin{itemize}
    \item For the grounding task, we enhance the dataset by applying horizontal flipping and random cropping to the images in \texttt{reg-grounding.jsonl} and \texttt{dist\_detect.jsonl}. For each sample, we construct augmented versions and integrate them into the original dataset at varying ratios of 15\%, 30\%, or 45\%.
    \item For the perception task, we focus on optimizing the \texttt{mcq.jsonl} dataset. By utilizing \texttt{train\_metadata.json}, we identify questions frequently appearing in the test set and generate a new version of \texttt{mcq.jsonl}. This new dataset then completely replaces the original \texttt{mcq.jsonl}.
    \item For the description task, our strategy is to increase the number of image quality levels. Specifically, we modify the \texttt{assess.jsonl}, \texttt{brief\_assess.jsonl}, and \texttt{scores.jsonl} datasets by replacing or augmenting the image quality levels within these files.
\end{itemize}

From a theoretical perspective, jointly training on heterogeneous yet semantically aligned data promotes shared representation learning across tasks, facilitating better generalization. Empirically, we observe that this augmentation-aware mixing not only maintains training efficiency but also consistently improves performance across grounding, perception, and description tasks, demonstrating its robustness and adaptability.

\subsection{Global Augmentation Strategies}
\label{sec_for_total}

\subsubsection{High-Resolution Input Enhancement}
To improve fine-grained perceptual sensitivity and task robustness across all three subtasks—quality grounding, perception, and description—we introduce a unified High-Resolution Input Enhancement strategy during fine-tuning. By increasing the resolution of input images, this strategy preserves more structural details and subtle distortion cues during training, providing richer visual information to support effective multi-task learning.

\subsubsection{Foundation Model Replacement}
To enhance the model’s ability to perceive and express fine-grained image quality information, we replace the base multimodal model with InternVL3~\cite{internvl3_doc}, which features stronger visual encoding and vision-language alignment capabilities. 
Compared to the Qwen2-VL~\cite{Qwen2VL}, InternVL3 provides better support for high-resolution inputs, localized region understanding, and multi-task instruction following—making it more suitable for IQA tasks that require both visual precision and interpretability. 

Our experiments confirm that this replacement leads to consistent performance gains across all subtasks. 
In this section, we fine-tune InternVL3 using the augmentation strategies introduced in Sections 3.1–3.3, along with the High-Resolution Input Enhancement strategy described below. This unified training pipeline enables the model to fully exploit the enriched data and better adapt to the multi-task setting of image quality grounding, perception, and description.

\section{Experiment Setting}

\subsection{Dataset}

\noindent\textbf{Training Set:}  
We use the ViDA-UGC dataset~\cite{DYEvalab2025ViDAMIPI} as our training set, consisting of 11,534 images annotated with Mean Opinion Scores (MOS) and an average of 3.6 distortion bounding boxes per image. ViDA-UGC incorporate GPT-generated degradation attributes to construct three task-specific subsets: ViDA-Description (for description), ViDA-Grounding (for grounding), and ViDA-Perception (for perception). Examples of images and annotations from the dataset are shown in Fig.~\ref{fig:intro}.

\noindent\textbf{Evaluation Set:}  
We adopt ViDA-UGC-Bench as the evaluation benchmark, comprising 476 images covering ten types of UGC distortions. The benchmark includes 476 description samples, 2,567 perception questions, and 3,106 distortion localization annotations. All test samples are manually verified to ensure quality and consistency.

\subsection{Models}
We fine-tune a variety of advanced MLLMs, including Qwen2-VL-7B-Instruct~\cite{Qwen2VL}, Qwen2.5-VL-7B-Instruct~\cite{qwen2.5vl}, InternVL3-8B-Instruct~\cite{internvl3}, and InternVL3-9B-Instruct~\cite{internvl3}, to comprehensively validate the effectiveness and generalization capability of our proposed methods.

\subsection{Evaluation Metrics}
We follow the ViDA-UGC-Bench protocol to evaluate models across three explainable IQA tasks:

\noindent\textbf{Grounding:} Mean Average Precision (mAP) is used to assess the model's ability to detect degradation regions. Both tasks are treated as multi-box detection problems, and predictions must include distortion types and normalized coordinates ([0,1000]).

\noindent\textbf{Perception:} Accuracy is adopted to evaluate the model’s performance on multiple-choice visual questions, measuring its understanding of distortion-related image content.

\noindent\textbf{Description:} Three metrics are used for evaluation: (1) Description (mAP) for all distortion detections, (2) Key Distortions Accuracy (ACC$_{0.5}$) for identifying key degradations, and (3) Image Quality Accuracy for predicting overall image quality levels (e.g., “fair”).

\subsection{Implementation Details}
We conduct our experiments using the ms-swift framework~\cite{ms-swift}, adopting Low-Rank Adaptation (LoRA)~\cite{lora} for fine-tuning with a rank of 16. Additionally, we unfreeze the Vision Transformer (ViT)~\cite{vit} and the cross-modal aligner to fully exploit cross-modal interactions. The training setup includes an initial learning rate of 4e-5, weight decay of 0.01, a warm-up ratio of 0.03, and cosine annealing for learning rate decay. All models are trained for a single epoch to ensure consistent and fair evaluation.

\section{Results and Analysis}

\subsection{Results on MIPI 2025 Challenge}
As shown in Table~\ref{tab_mipi}, our method achieves state-of-the-art performance across all three sub-tasks—Grounding, Perception, and Description—when compared with other methods on the current leaderboard. This consistently strong performance demonstrates the effectiveness of our task-specific augmentation strategies. In particular, the High-Resolution Input Enhancement significantly improves the model’s ability to perceive fine-grained distortion patterns.

From a theoretical perspective, joint training on heterogeneous multi-source data enhances the model's capacity to learn shared representations across tasks, thereby improving overall multi-task learning effectiveness. Experimental results confirm that our proposed mixed-data fine-tuning scheme yields consistent gains across all sub-tasks, highlighting the model’s strong adaptability and stability.
\begin{table*}[t]
\begin{center}
\caption{Results of the MIPI 2025 Detailed Image Quality Assessment Challenge.}
\label{tab_mipi}
\vspace{-8pt}
{\begin{tabular}{@{}ccccccccc@{}}
\toprule[1.0pt]
Rank & \begin{tabular}[c]{@{}c@{}}Team \\ Name \end{tabular}          & \begin{tabular}[c]{@{}c@{}}Final \\ Score \end{tabular}    & \begin{tabular}[c]{@{}c@{}}Perception \\ Accuracy \end{tabular}   & \begin{tabular}[c]{@{}c@{}}Region \\ mAP \end{tabular}     & \begin{tabular}[c]{@{}c@{}}Distortion \\ mAP \end{tabular} & \begin{tabular}[c]{@{}c@{}}Description \\ mAP \end{tabular} & \begin{tabular}[c]{@{}c@{}}Key Distortion \\ Accuracy \end{tabular}   & \begin{tabular}[c]{@{}c@{}}Image Quality \\ Accuracy \end{tabular} \\ \midrule
1    & IH-VQA (ours)             & \textbf{2.80} & \textbf{0.81}       & \textbf{0.40} & \textbf{0.12}  & \textbf{0.20}    & \textbf{0.43}           & \textbf{0.83}          \\
2    & CrazyCat           & 2.43         & 0.72                & 0.33         & 0.11           & 0.15            & 0.37                    & 0.76                   \\
3    & MediaX             & 2.43         & 0.72                & 0.33         & 0.11           & 0.15            & 0.37                    & 0.76                   \\
4    & Smart vision group & 2.26         & 0.72                & 0.14         & 0.11           & 0.14            & 0.37                    & 0.78                   \\
5    & IVP-Lab            & 1.67         & 0.52                & 0.08         & 0.02           & 0.05            & 0.38                    & 0.61                   \\
6    & echoch             & 0.70          & 0.70                 & 0.00            & 0.00              & 0.00               & 0.00                       & 0.00                      \\ \bottomrule[1.0pt]
\end{tabular}}
\end{center}
\vspace{-4pt}
\end{table*}

\subsection{Ablation Study}
To validate the effectiveness of the proposed augmentation strategies, we conduct a series of systematic ablation studies. Sections~\ref{sec:ablation_grounding}–\ref{sec:ablation_description} analyze the task-specific augmentation methods designed for Grounding, Perception, and Description, respectively, and examine their individual contributions to performance improvements. Furthermore, Section~\ref{sec:ablation_total} evaluates the impact of the High-Resolution Input Enhancement strategy and backbone selection, offering a comprehensive understanding of how each component influences overall system performance.

\subsubsection{Ablations and Analysis for Grounding}
\label{sec:ablation_grounding}

{
Table \ref{grounding_aug_ablation} shows the results of an ablation study on different augmentation methods. The performance drop from individual augmentations suggests they can disrupt spatial cues. The success of their combination is due to their complementary roles: Horizontal Flipping generalizes the appearance of distortions, while Random Cropping generalizes their context. This forces the model to learn more robust, location-agnostic features, thus improving overall grounding performance.
}

\begin{table}[t]
\begin{center}
\caption{Performance comparison of different augmentation methods for the grounding task.}
\label{grounding_aug_ablation}
\vspace{-8pt}
\setlength{\tabcolsep}{1.7mm}{\begin{tabular}{@{}cccc@{}}
\toprule[1.0pt]
Method                        & Region mAP & Distortion mAP & Avg.            \\ \midrule
Original                      & \textbf{0.3899}     & 0.1083         & 0.2491          \\
Horizontal Flip.               & 0.3693     & \textbf{0.1232}         & 0.2463          \\
Random Crop.                   & 0.3615     & 0.1086         & 0.2351          \\
\begin{tabular}[c]{@{}c@{}}Horizontal Flip. \\ + Random Crop.\end{tabular} & 0.3892     & 0.1210         & \textbf{0.2551} \\ \bottomrule[1.0pt]
\end{tabular}}
\end{center}
\vspace{-8pt}
\end{table}

Table \ref{grounding_flip_ablation} shows the results of applying different percentages of Horizontal Flipping augmentation. The baseline model (Original) performs the worst, while adding 15\% Horizontal Flipping slightly improves the performance. Increasing the augmentation to 30\% results in a small drop, but 45\% Horizontal Flipping gives the best performance, with the highest average score of 0.2518.

\begin{table}[t]
\begin{center}
\caption{Impact of horizontal flipping augmentation percentage on the grounding task.}
\label{grounding_flip_ablation}
\vspace{-8pt}
\setlength{\tabcolsep}{0.2mm}{\begin{tabular}{@{}cccc@{}}
\toprule[1.0pt]
Method               & Region mAP & Distortion mAP & Avg.            \\ \midrule
Original             & \textbf{0.3899}     & 0.1083         & 0.2491          \\
15\% Horizontal Flip. & 0.3773     & 0.1215         & 0.2494          \\
30\% Horizontal Flip. & 0.3693     & \textbf{0.1232}         & 0.2463          \\
45\% Horizontal Flip. & 0.3833     & 0.1203         & \textbf{0.2518} \\ \bottomrule[1.0pt]
\end{tabular}}
\end{center}
\vspace{-8pt}
\end{table}

\subsubsection{Ablations and Analysis for Perception}
\label{sec:ablation_perception}

Table \ref{ablation_mcq} compares the accuracy of different augmentation methods for the perception task. The Self-made method achieves the highest accuracy of 0.7620, significantly improving over the Original method, which has an accuracy of 0.5462. The Shuffle Options method results in a slight decrease to 0.7616, while the More Options method leads to an accuracy of 0.7437. The Self-made method is the most effective in addressing the issue of inconsistent question distributions between the training and test sets.

\begin{table}[t]
\begin{center}
\caption{Accuracy comparison of different augmentation methods using Qwen2-VL-7B-Instruct for the perception task.}
\label{ablation_mcq}
\vspace{-8pt}
\begin{tabular}{@{}cc@{}}
\toprule[1.0pt]
Method          & Perception Accuracy \\ \midrule
Original        & 0.5462              \\
Self-made       & \textbf{0.7620}     \\
Shuffle Options & 0.7616              \\
More Options    & 0.7437              \\ \bottomrule[1.0pt]
\end{tabular}
\end{center}
\vspace{-12pt}
\end{table}

\subsubsection{Ablations and Analysis for Description}
\label{sec:ablation_description}

Table \ref{ablation_score} shows the impact of varying the image quality level granularity on the image quality accuracy for the description task. Using 5 levels results in an accuracy of 0.7857, while increasing the levels to 10 improves the accuracy to 0.8046, the highest among all configurations. With 15 levels, the accuracy slightly drops to 0.7983, and with 20 levels, it increases again to 0.8004. Overall, the results suggest that finer granularity, particularly with 10 levels, leads to better performance in predicting image quality.

\begin{table}[t]
\begin{center}
\caption{Impact of image quality level granularity using Qwen2-VL-7B-Instruct for the description task.}
\label{ablation_score}
\vspace{-8pt}
\begin{tabular}{@{}cc@{}}
\toprule[1.0pt]
Image Quality Level & Image Quality Accuracy \\ \midrule
5     & 0.7857                 \\
10    & \textbf{0.8046}        \\
15    & 0.7983                 \\
20    & 0.8004                 \\ \bottomrule[1.0pt]
\end{tabular}
\end{center}
\vspace{-4pt}
\end{table}

\subsubsection{Ablations for High-Resolution Input Enhancement and Foundation Model }
\label{sec:ablation_total}
\paragraph{High-Resolution Input Enhancement:} 
{
We evaluated the impact of input image resolution, controlled by the number of max pixel tokens, across all three tasks (Tables \ref{ablation_pixel_grounding}, \ref{ablation_pixel_perception}, and \ref{ablation_pixel_score}). A clear and consistent trend was observed: model performance generally improves as the input resolution increases. Specifically, in the grounding task, performance showed a steady and significant improvement as the resolution increased, reaching its highest metric score when using 2048 maximum pixel tokens. For the perception task, a similar upward trend is observed, although there is a slight performance dip at 512 max pixel tokens before improving again at higher resolutions. In the description task, accuracy increases with resolution, reaching its peak at 2048 max pixel tokens. Across all three tasks, the optimal performance was consistently achieved at 2048 max pixel tokens, demonstrating that this resolution provides the best balance of fine-grained detail and contextual information for comprehensive image quality assessment.
}

\begin{table}[t]
\begin{center}
\caption{Impact of resolution using Qwen2-VL-7B-Instruct for the grounding task.}
\label{ablation_pixel_grounding}
\vspace{-8pt}
\setlength{\tabcolsep}{0.8mm}{
\begin{tabular}{@{}cccc@{}}
\toprule[1.0pt]
Max Pixel Tokens & Region mAP      & Distortion mAP  & Avg.            \\ \midrule
256        & 0.3099          & 0.0923          & 0.2011          \\
512        & 0.3427          & 0.0979          & 0.2203          \\
1024       & 0.3445          & \textbf{0.1162} & 0.2304          \\
2048       & \textbf{0.3658} & 0.1055          & \textbf{0.2356} \\ \bottomrule[1.0pt]
\end{tabular}}
\end{center}
\vspace{-4pt}
\end{table}

\begin{table}[t]
\begin{center}
\caption{Impact of resolution with more options using Qwen2-VL-7B-Instruct for the perception task.}
\label{ablation_pixel_perception}
\vspace{-8pt}
\begin{tabular}{@{}cc@{}}
\toprule[1.0pt]
Max Pixel Tokens & Perception Accuracy \\ \midrule
256        & 0.7616              \\
512        & 0.7503              \\
1024       & 0.7628              \\
2048       & \textbf{0.7643}     \\ \bottomrule[1.0pt]
\end{tabular}
\end{center}
\vspace{-4pt}
\end{table}

{
Our analysis shows the High-Resolution Input Enhancement module's impact is task-specific. It provides a significant boost to the grounding task, which requires the fine-grained spatial details preserved in high-resolution inputs, while its influence on the more global perception and description tasks is less pronounced.
}

\begin{table}[t]
\begin{center}
\caption{Impact of resolution with 10 levels image quality using Qwen2-VL-7B-Instruct for the description task.}
\label{ablation_pixel_score}
\vspace{-8pt}
\begin{tabular}{@{}cc@{}}
\toprule[1.0pt]
Max Pixel Tokens & Image Quality Accuracy \\ \midrule
256        & 0.8046                 \\
512        & 0.8025                 \\
1024       & 0.8151                 \\
2048       & \textbf{0.8172}        \\
4096       & 0.8151                 \\ \bottomrule[1.0pt]
\end{tabular}
\end{center}
\vspace{-4pt}
\end{table}

\vspace{-8pt}
\paragraph{Foundation Model:} 

{
The ablation study compares various foundation models and their performance across tasks. The Qwen2-VL-7B-Instruct model has the lowest performance in both grounding and perception tasks, with an average score of 0.2551 and accuracy of 0.7620, respectively. The Qwen2.5-VL-7B-Instruct model shows slight improvements in both areas, achieving an average score of 0.2707 and accuracy of 0.7709. InternVL3-8B-Instruct performs the best, with the highest grounding score (0.2768) and accuracy in perception (0.7982) and image quality (0.8151) tasks. The InternVL3-9B-Instruct model, while still strong, underperforms compared to InternVL3-8B, especially in grounding and image quality. These results highlight the superior performance of InternVL3-8B-Instruct across various tasks.
}

\begin{table}[t]
\begin{center}
\caption{Impact of foundation models with horizontal flipping and random cropping augmentation for the grounding task.}
\label{ablation_model_grounding}
\vspace{-8pt}
{\begin{tabular}{@{}cccc@{}}
\toprule[1.0pt]
Model & \begin{tabular}[c]{@{}c@{}}Region \\ mAP \end{tabular} & \begin{tabular}[c]{@{}c@{}}Distortion \\ mAP \end{tabular} & Avg.            \\ \midrule
Qwen2-VL-7B-Instruct     & 0.3892     & 0.1210         & 0.2551          \\
Qwen2.5-VL-7B-Instruct   & 0.4115     & 0.1299         & 0.2707          \\
InternVL3-8B-Instruct & \textbf{0.4220}     & \textbf{0.1317}         & \textbf{0.2768} \\
InternVL3-9B-Instruct & 0.3920     & 0.1221         & 0.2570          \\ \bottomrule[1.0pt]
\end{tabular}}
\end{center}
\vspace{-16pt}
\end{table}

\begin{table}[t]
\begin{center}
\caption{Impact of foundation models with self-made augmentation for the perception task.}
\label{ablation_model_mcq}
\vspace{-8pt}
\setlength{\tabcolsep}{0.7mm}{\begin{tabular}{@{}cc@{}}
\toprule[1.0pt]
Model                 & Perception Accuracy \\ \midrule
Qwen2-VL-Instruct     & 0.7620              \\
Qwen2.5-VL-Instruct   & 0.7709              \\
InternVL3-8B-Instruct & \textbf{0.7982}     \\
InternVL3-9B-Instruct & 0.7709              \\ \bottomrule[1.0pt]
\end{tabular}}
\end{center}
\vspace{-8pt}
\end{table}

\begin{table}[t]
\begin{center}
\caption{Impact of foundation models with 10 levels image quality for the description task.}
\label{ablation_model_score}
{\begin{tabular}{@{}cc@{}}
\toprule[1.0pt]
Model                 & Image Quality Accuracy \\ \midrule
Qwen2-VL-Instruct     & 0.8046                 \\
Qwen2.5-VL-Instruct   & 0.7920                 \\
InternVL3-8B-Instruct & \textbf{0.8151}        \\
InternVL3-9B-Instruct & 0.7857                 \\ \bottomrule[1.0pt]
\end{tabular}}
\end{center}
\vspace{-14pt}
\end{table}
\section{Conclusion and Future Work}

We present iDETEX, a unified MLLM-based network tailored for detailed and explainable image quality assessment. By jointly addressing quality grounding, perception, and description, iDETEX moves beyond traditional black-box IQA scoring. Built on three task-specific data augmentation modules and a Task-Aware Augmented Data Mixing Strategy, our framework enables efficient and generalizable multi-task learning. Furthermore, online input enhancement improves robustness under diverse distortion scenarios. On the ViDA-UGC benchmark, iDETEX consistently achieves SOTA performance, leading the leaderboard of the MIPI 2025 Detailed Image Quality Assessment Challenge. These results highlight our paradigm’s potential to advance interpretable, fine-grained IQA leveraging MLLMs.

iDETEX is trained on large-scale ViDA-UGC dataset to ensure robust performance, raising considerations for low-data scenarios. We posit primary path to scalability is transfer learning. The pre-trained model on this diverse dataset serves as a powerful backbone, enabling efficient fine-tuning on smaller, specialized datasets (e.g., niche distortion types) with less annotation. Validating this transferability and exploring complementary few-shot learning techniques are promising future research directions.
\clearpage
{
    \small
    \bibliographystyle{ieeenat_fullname}
    \bibliography{main}
}


\end{document}